\documentclass[11pt]{article}

\usepackage[margin=1in]{geometry}
\usepackage{amsmath,amssymb,amsthm}
\usepackage{booktabs}
\usepackage{graphicx}
\usepackage{hyperref}
\usepackage{natbib}
\usepackage{enumitem}
\usepackage{xcolor}
\usepackage{microtype}
\usepackage{caption}
\usepackage{subcaption}
\usepackage{multirow}
\usepackage{array}

\hypersetup{colorlinks=true, linkcolor=blue!60!black, citecolor=blue!60!black, urlcolor=blue!60!black}

\newtheorem{definition}{Definition}[section]
\newtheorem{proposition}[definition]{Proposition}

\title{\textbf{Theoria: Rewrite-Acceptability Verification\\ over Informal Reasoning
States}\footnote{A preliminary white paper describing the framework, architecture, and the primary HLE-Verified Gold evaluation accompanied the project's June 2026 open-source release: \texttt{github.com/zaladbar/theoria}.}}

\author{
  Michael Saldivar\thanks{Corresponding author: \texttt{msaldi@alum.mit.edu}} \quad Ben Slivinski \\
  Independent Researchers \\
  \texttt{msaldi@alum.mit.edu} \\
  \texttt{github.com/zaladbar/theoria}
}

\date{}

\begin{document}
\maketitle

\begin{abstract}
When should an AI system's answer be trusted? Formal proof assistants offer certainty but cannot reach most of the problem distribution; scalar LLM judges offer coverage but produce opaque scores that cannot be audited after the fact and are subject to the same coherence issues as any LLM. We present Theoria, a verification architecture that closes this gap. A candidate solution is rewritten into a sequence of typed state transitions, each licensed by an explicit justification, whether that be a citation, computation, or problem-given fact, and every transition is independently auditable. The foundational invariant is \emph{completeness of change}: every difference between consecutive proof states must be accounted for, so hidden premises surface as unlicensed mutations rather than passing silently.

On HLE-Verified Gold (185 text-only expert problems), Theoria certifies 105 at 91.4\% strict precision (Wilson 95\% CI [84.5\%, 95.4\%]). Every certification produces a human-readable proof trace in which each step can be independently challenged. Holistic LLM judges achieve comparable precision at matched coverage but fail on different problems (Jaccard 0.14--0.36), making the approaches complementary. On 95 adversarial poisoned proofs across 15 domains, structured judges catch 94.7\% versus 83.2\% for holistic judging ($p = 0.0017$). The overall 11.5~pp gap concentrates in hidden premises (90.6\% vs.\ 62.5\%, a 28~pp difference) and fabricated citations (100\% vs.\ 90\%), the error classes where the formal analysis predicts an advantage; performance is identical on arithmetic and theorem-misapplication errors, where no advantage is predicted. On GPQA Diamond ($n=65$), certified precision is 97.1\% (Wilson CI [85.1\%, 99.5\%]).

\end{abstract}

\section{Introduction}

Reasoning verification is not the problem of making an AI system produce more correct answers on average. It is the problem of deciding when a produced answer should create justified reliance. For casual use, a plausible answer suffices. For scientific discovery, legal analysis, financial modeling, medical decision support, engineering verification, or any other "safety critical field", the relevant object is not the answer alone but the steps by which the system proposes to make that answer usable. A verifier that merely improves average accuracy has not solved the trust problem. A verifier must draw a boundary: these outputs have survived a procedure with a known failure surface, those outputs have not.

Today, two approaches exist. At one end, formal methods make correctness a kernel property. Proof assistants like Lean~\citep{lean}, Coq/Rocq~\citep{rocq}, and Isabelle~\citep{nipkow2002isabelle} accept or reject proof terms, and modern AI systems exploit this by translating natural-language problems into formal targets and searching for formal proofs~\citep{deepseek_prover, alphaproof_blog, alphaproof_nature, alphaproof_nexus, aristotle, axiomprover}. Formal verification yields strong results once the specification is fully formalized. However, this approach becomes fragile at the point where informal, natural-language requirements must be interpreted. When the primary challenge lies in determining which formal specification accurately reflects the real-world problem, the verification kernel may flawlessly prove an incorrect theorem. Hence, the barrier to autoformalization is not just a limitation of current tooling, but a fundamental semantic boundary.

At the other end, scalar reward and judge systems operate over ordinary reasoning traces and cover far more of the problem distribution. Process reward models~\citep{lightman2023lets, math_shepherd, implicit_prm} provide dense training signal and LLM-as-judge pipelines~\citep{zheng2023judging, gu2024survey} scale qualitative evaluation. Their weakness is that a score is not a "verification certificate". A scalar can be right or wrong, but it cannot be complete or incomplete in the sense that a proof can. LLM-as-judge evaluations, relying on qualitative reasoning, inherit these same logical vulnerabilities. It does not record which premise was used, which transformation was licensed, or whether an additional assumption entered silently.

Theoria occupies a third point in this space. It does not translate informal reasoning into Lean nor does it ask an LLM to assign a holistic quality score. Instead, it rewrites the solver's answer into a witness, that is, an initial state and a sequence of state-to-state transformations. Each transformation carries exactly one justification type: \emph{citation}, \emph{computation}, or \emph{problem\_given}. A verifier asks whether the stated justification licenses the entire change from the previous state to the next. This is a local question, and it is a different question from ``does the conclusion follow?'' The rewrite question is: given this before state, this after state, this justification type, and this evidence, is the observed change acceptable?

The idea is that a well-chosen witness format can make verifier fallibility less destructive. LLM judges are not trusted as oracles. They are assigned constrained local obligations against an explicit diff. The proof witness is designed so that the errors most likely to matter, namely assumptions, fabricated citations, unjustified computations, silent convention choices, must surface as visible changes in the state, where an adversarial judge can catch them. The architecture no longer depends on the judge reconstructing an entire proof from prose. This is a concrete instantiation of \emph{scalable oversight}: the system produces structured work products that humans or downstream auditors can inspect at tractable granularity, without reconstructing latent reasoning from an ambiguous output.

Sections~\ref{sec:related}--\ref{sec:framing} present related work and the rewrite framing. Section~\ref{sec:formal} states the formal properties of rewrite witnesses, including the exposure guarantee for hidden premises and testable predictions about which error classes benefit. Sections~\ref{sec:architecture}--\ref{sec:structural} describe the architecture and the structural argument. Section~\ref{sec:calibration} discusses calibration. Section~\ref{sec:experiments} reports the empirical evaluation on HLE-Verified Gold, including a solver-only baseline (Section~\ref{sec:solver_baseline}), holistic-judge baselines (Section~\ref{sec:holistic}), an error-overlap analysis (Section~\ref{sec:overlap}), cross-model adjudication (Section~\ref{sec:adjudication}), an adversarial robustness test that directly tests the predictions of Proposition~\ref{prop:error_classes} (Section~\ref{sec:adversarial}), and an out-of-distribution evaluation on GPQA Diamond (Section~\ref{sec:gpqa}).

\section{Related Work}
\label{sec:related}

Theoria sits at the intersection of three lines of prior work. The contribution is not that it uses multiple models, step-checking, or repair (these ideas are fairly common). The contribution is the specific witness format: state rewrites over informal reasoning, with typed justifications and a completeness-of-change invariant.

\paragraph{Formal proof assistants and autoformalization.}
Mizar~\citep{mizar}, Coq/Rocq~\citep{rocq}, Isabelle/HOL~\citep{nipkow2002isabelle}, and Lean~\citep{lean} provide trusted kernels where a proof term either checks or does not. AI autoformalization attacks the entry cost. DeepSeek-Prover~\citep{deepseek_prover} uses large-scale synthetic Lean data. AlphaProof~\citep{alphaproof_blog, alphaproof_nature} combines learned search with Lean verification at Olympiad level. AlphaProof Nexus~\citep{alphaproof_nexus}, Aristotle~\citep{aristotle}, and AxiomProver~\citep{axiomprover} extend the formal track. These systems help clarify Theoria's boundary. Theoria is designed for the space where the problem remains informal, the correct formal target is not obvious, or formalization cost exceeds the use case. Because a Theoria witness already exposes each premise and state change under a typed local license, it is a more tractable autoformalization target than free-form prose. In theory, a formal prover can be attached downstream to complete formalizable steps.

\paragraph{Process supervision and learned verification.}
Lightman et al.~\citep{lightman2023lets} showed process supervision outperforms outcome supervision for mathematical reasoning. Math-Shepherd~\citep{math_shepherd} constructs process-wise supervision automatically. Implicit PRMs~\citep{implicit_prm} extract step-level signals from outcome-level data. ReST~\citep{rest} and ReST-MCTS*~\citep{rest_mcts} use reward signals to guide reasoning generation. These systems treat the step as an object of evaluation. The difference is the type of object returned. A PRM outputs a scalar that can guide sampling, reranking, or reinforcement learning, but it does not track the logical steps between premises, nor does it identify every state change. Scalar reward is what policy optimization needs, but it is not a substitute for a proof witness.

\paragraph{LLM-as-judge.}
Zheng et al.~\citep{zheng2023judging} established LLM-as-judge as a scalable evaluation method. The literature has documented position effects, verbosity effects, self-enhancement, and prompt sensitivity~\citep{gu2024survey}. Theoria uses LLM judges but gives them a narrower role: not rating global correctness, but answering a typed local question about whether a cited justification licenses a specific state change.

\paragraph{Structured natural-language verification.}
Ling et al.~\citep{ling2023deductive} decompose reasoning verification into step-by-step subprocesses. MATH-VF~\citep{math_vf} uses a formalizer and critic with CAS tools. Prover-verifier games~\citep{kirchner2024prover} optimize for legibility to smaller verifiers. SAVeR~\citep{saver} and graph-structured verification~\citep{graph_verification} model intermediate belief states. These works share Theoria's broad intuition that the format of reasoning is part of verification. Theoria differs in two ways: (1) it tracks changes to states rather than logical steps between statements, and (2) it uses few operational tags, like citation, computation, or problem given, to keep instructions simple and vocabulary unambiguous.

\section{The Rewrite Framing}
\label{sec:framing}

\subsection{Proof Witnesses as State Trajectories}

A Theoria proof witness consists of an initial state $S_0$ and a finite sequence of steps
{\setlength{\abovedisplayskip}{3pt}
\setlength{\belowdisplayskip}{3pt}
\[
(S_1, \tau_1, e_1), (S_2, \tau_2, e_2), \dots, (S_n, \tau_n, e_n)
\]}

Each state $S_i$ is an ordered list of strings representing what is currently established and what remains to be determined. By convention, index~0 holds the expression being solved for: initially a goal variable (e.g., ``ANSWER~=~?'') and finally the resolved answer. This representation is intentionally less formal than the core language of a proof assistant and more constrained than the free prose. It is not a proof term, but instead a state object over which differences can be inspected.

A step $(S_i, \tau_i, e_i)$ is a transition from state $S_{i-1}$ to state $S_i$, justified by evidence $e_i$ of type $\tau_i \in \{\texttt{citation}, \texttt{computation}, \texttt{problem\_given}\}$. The verifier receives the problem statement and the full proof as context, then asks a deliberately local question: does justification $e_i$ of type $\tau_i$ license the entire change from $S_{i-1}$ to $S_i$?

\subsection{The Completeness-of-Change Invariant}

Let $\Delta(S_{i-1}, S_i)$ denote the substantive changes introduced by a step: added claims, removed claims, strengthened hypotheses, instantiated parameters, rewritten expressions, changed domains, altered definitions, and changes to the answer slot. Let $L_\tau(e;\, S_{i-1}, P)$ denote the set of changes licensed by evidence $e$ of type $\tau$ in the context of previous state $S_{i-1}$ and problem $P$. The rewrite-acceptability condition is:
\begin{equation}
\Delta(S_{i-1},\, S_i) \;\subseteq\; L_\tau(e_i;\, S_{i-1},\, P)
\label{eq:coc}
\end{equation}
with the additional requirement that all premises needed to apply the license are present in $S_{i-1}$ or were introduced by a prior justified step.

This is the \emph{completeness-of-change invariant}. Every change must be accounted for. A proof step is not acceptable merely because the new state is true or globally derivable from the old state. It is acceptable only if the stated justification accounts for the observed transformation. The invariant is a conservation law over proof-state changes. It prevents premises from entering silently. In ordinary prose, a proof can move from ``let $p$ be an odd prime'' to ``therefore the Legendre symbol has property $X$'' while silently importing $p \nmid a$, a convention about least residues, and a theorem whose hypotheses are only partly met. In a rewrite witness, each imported condition must either be present in the prior state or appear as a new element of $S_i$ (where a judge can ask whether the citation licenses it). 

\subsection{Justification Types}

\texttt{Citation} covers theorems, identities, definitions, laws, and rules of inference. The cited result must exist, apply under the current state, and account for all changes. It cannot license unstated side conditions or conclusions stronger than the theorem supplies.

\texttt{Computation} covers arithmetic, algebraic simplification, symbolic manipulation, enumeration, and deterministic operations. Its obligation is primarily correctness and proper inclusion. A computation can license $2x + 3 = 7$ becoming $x = 2$. It cannot silently assume $x$ is real if the domain matters, or divide by a quantity without establishing it is nonzero.

\texttt{Problem\_given} is the strictest type. It imports only facts directly stated in the problem text or notation. If a problem says ``let $p$ be prime,'' the state may import ``$p$ is prime.'' But if a problem \emph{implies} a consequence, that implication requires a citation and possibly a computation (it is not problem) given. The purpose is to prevent the system from placing convenient facts in early states.

\subsection{State~0 and the Initial-State Audit}

Since the initial state has no predecessor and no justification field, it is a natural place for unjustified premises. Theoria audits $S_0$ independently. The target answer slot must be a goal, not a conclusion. Every other entry must be present in the problem text or a notational restatement requiring no reasoning. 

\subsection{Relationship to Prior Proof Frameworks}

The connection to Hoare logic~\citep{hoare1969axiomatic} is worth noting. A Hoare triple $\{P\}\, C\, \{Q\}$ says that command $C$, run from a state satisfying $P$, terminates in a state satisfying $Q$. Theoria applies the same intuition to informal reasoning. The analogy is not perfect (i.e natural-language states are not program stores, and LLM judges are not proof rules) but the design inspiration is relatively clear.

As Razborov and Rudich~\citep{razborov1997natural} showed for circuit lower bounds, the choice of proof format determines which failures are structurally visible and which remain exploitable. This is a lesson that motivates Theoria's insistence on explicit state transitions over opaque, ambiguous chain-of-thought reasoning.

The connection to interactive proof systems~\citep{babai1988arthur, goldwasser1986private} is also worth discussing. In Arthur--Merlin's protocol, a prover constructs a witness and a verifier checks it with bounded error. Theoria's verifier is an LLM ensemble, not a probabilistic Turing machine, and its witnesses are informal. But the design question is similar: what witness format makes the verifier's job tractable?

\section{Formal Properties of Rewrite Witnesses}
\label{sec:formal}

The preceding section introduced the rewrite framing and its notation. This section states the core properties precisely, distinguishes two failure modes, and derives predictions that the experiments in Section~\ref{sec:experiments} test directly.

\subsection{Definitions}

We build on the notation of Section~\ref{sec:framing}. A \emph{proof witness} is a tuple $(P,\, S_0,\, \{(S_i, \tau_i, e_i)\}_{i=1}^n)$ where $P$ is the problem statement, $S_0$ is the initial state, and each step transitions from $S_{i-1}$ to $S_i$ with evidence $e_i$ of type $\tau_i$.

\begin{definition}[State mutation]
\label{def:mutation}
The \emph{mutation set} $\Delta(S_{i-1}, S_i)$ is the set of semantic changes between consecutive states: claims added, removed, strengthened, instantiated, or rewritten. A mutation $\delta \in \Delta(S_{i-1}, S_i)$ is \emph{licensed} if $\delta \in L_\tau(e_i;\, S_{i-1}, P)$ and \emph{unlicensed} otherwise.
\end{definition}

\begin{definition}[Hidden premise]
\label{def:hidden}
A \emph{hidden premise} in step $i$ is a claim $c$ required to derive $S_i$ from $S_{i-1}$ that is neither present in $S_{i-1}$ nor entailed by evidence $e_i$ of type $\tau_i$.
\end{definition}

\begin{definition}[Exposure failure and judge failure]
\label{def:failure_modes}
An error in a proof can go undetected in two distinct ways:
\begin{enumerate}[nosep]
\item \textbf{Exposure failure}: the witness format does not represent the error as an observable feature that a correct verifier would reject.
\item \textbf{Judge failure}: the witness exposes the error but the judge fails to detect it.
\end{enumerate}
The total miss rate decomposes as:
\[
P(\text{miss}) = P(\text{not exposed}) + P(\text{exposed}) \cdot P(\text{judge miss} \mid \text{exposed})
\]
Theoria's theoretical claim is that the rewrite format reduces the first term. The second term depends on judge capability and is addressed empirically, not architecturally.
\end{definition}

\subsection{Completeness of Change as an Exposure Guarantee}

\begin{proposition}[Hidden premises produce unlicensed mutations]
\label{prop:exposure}
Let $(P, S_0, \{(S_i, \tau_i, e_i)\}_{i=1}^n)$ be a proof witness satisfying the completeness-of-change invariant (Equation~\ref{eq:coc}): every semantic difference between $S_{i-1}$ and $S_i$ appears in $\Delta(S_{i-1}, S_i)$. If step $i$ depends on a hidden premise $c$, then either:
\begin{enumerate}[nosep]
\item[\emph{(a)}] $c$ appears as a new element of $S_i$, producing a mutation $\delta_c \in \Delta(S_{i-1}, S_i)$ that is not licensed by $e_i$ (since $c$ is not entailed by the stated evidence), or
\item[\emph{(b)}] $c$ does not appear in $S_i$, violating completeness of change (the derivation depends on $c$ but $c$ is absent from the state).
\end{enumerate}
In either case, a correct verifier rejects step $i$ (a) for an unlicensed mutation, or (b) for an incomplete state representation.
\end{proposition}

The consequence is that a global reasoning gap, an unstated assumption somewhere in a multi-step argument, is converted into a local, diff-level observable. In this way, the judge does not need to reconstruct the full argument to notice the gap.

\subsection{Differential Exposure by Error Class}

Not all error types benefit equally from the rewrite format.

\begin{proposition}[Error-class predictions]
\label{prop:error_classes}
Under a rewrite witness with completeness of change:
\begin{enumerate}[nosep]
\item \textbf{Hidden premises} are exposed by Proposition~\ref{prop:exposure}: they must appear as unlicensed mutations.
\item \textbf{Fabricated citations} are exposed because evidence $e_i$ is explicitly presented with its type label; a citation judge can verify existence and applicability against a specific state diff.
\item \textbf{Arithmetic errors} receive no \emph{exposure} advantage: the error is a wrong number, which is visible in both prose and a state transition (unlike a hidden premise, which is invisible without the diff). Step isolation may reduce attentional load on the judge, but detection depends on judge capability rather than format-created visibility.
\item \textbf{Misapplied theorems} receive no \emph{exposure} advantage: the cited theorem is real and the error lies in whether its hypotheses are met, a semantic judgment required in either format. Step isolation narrows the context the judge must consider, but does not create visibility that would otherwise be absent.
\end{enumerate}
\end{proposition}

This yields a testable prediction: the detection-rate advantage of structured over holistic verification should concentrate in hidden premises and fabricated citations, with little or no advantage on arithmetic errors and theorem misapplication. Section~\ref{sec:adversarial} tests this prediction on 95 poisoned proofs.

\subsection{Limits of Witness-Based Verification}

\begin{proposition}[Unobservable errors]
\label{prop:impossibility}
No witness-based verifier can detect an error whose semantic effect produces no observable state mutation. If a reasoning error leaves all states $S_0, \ldots, S_n$ unchanged and all license checks $\Delta(S_{i-1}, S_i) \subseteq L_\tau(e_i;\, S_{i-1}, P)$ satisfied, the error is invisible to any verifier operating on the witness alone.
\end{proposition}

Two concrete examples: an initial state $S_0$ that subtly misinterprets the problem (e.g., reading ``at most'' as ``at least''), or a sequence of individually licensed steps that silently shifts the meaning of a term across steps without any single step producing a detectable mutation. The initial-state judge (Section~\ref{sec:architecture}) mitigates the first case by checking $S_0$ against the problem text, but subtle misinterpretations can survive this check which is a limitation shared by formal-verification systems such as Lean, where the gap between a natural-language problem and its formal specification is a well-known source of error. These are errors of \emph{interpretation}, not of \emph{derivation}, and they define the boundary of what witness-based verification can guarantee.

\section{Architecture}
\label{sec:architecture}

\subsection{Overview}

The Theoria pipeline runs the following loop: solve $\to$ formalize $\to$ judge $\to$ filter $\to$ repair. A solver proposes a natural-language answer. A formalizer converts it into a proof witness: an initial state and a sequence of typed steps. Specialized judges audit the witness in parallel. If all audits pass, the system returns \textsc{judge-passed}. If any fails, the failure passes through a pedantry filter and an optional convention lift before triggering a bounded repair loop or a final decline.

Neither the solver nor the formalizer is trusted. The solver proposes and the formalizer rewrites. A solver can be globally goal-directed and creative. A formalizer must be locally conservative and explicit. When the formalizer discovers that the solver used a false fact or made an unjustified leap, it can reject the solution before any judge calls are spent.

\subsection{Witness Construction}

The formalizer's important constraint is that premises must be explicit before use. If a step relies on a bound, condition, or assumption not already in the previous state, the proof must first add that premise in its own justified step. This, simply put, is the operational version of completeness-of-change. 

\subsection{Parallel Specialized Judging}

Judging is parallelized by step and by justification type. A computation step goes to a computation judge that recomputes from scratch and checks for hidden assumptions. A citation step goes to a citation judge that verifies existence and correct application. A problem\_given step goes to a judge that checks the problem text directly. State~0 goes to an independent initial-state judge. These roles are specialized by design. The point is to reduce each local audit to the right kind of skepticism. All judges are adversarial in prompt design (i.e told that LLMs hallucinate and fabricate, and asked to find errors rather than confirm plausibility). 

\subsection{Pedantry Filter}

Natural-language mathematical problems contain conventions, equivalent notations, and informal compression. A judge demanding maximal rigor may reject a step that is acceptable at the problem's intended standard. The pedantry filter examines each rejection and asks whether it points to a substantive error or merely to over-strictness about cosmetic compression.

Judges must exhaustively enumerate every issue as separately numbered points. The pedantry filter then requires that \emph{every} enumerated issue be pedantic before the override fires. A single substantive issue blocks the override. This prevents a multi-issue rejection from being overridden because one cosmetic issue appears alongside a substantive one.

\subsection{Convention Lift}

After pedantry, a legitimate rejection may still be resolvable by invoking a single standard, citable, domain-level convention. The convention must satisfy three criteria: (1) practitioners would assume it by default; (2) it has a verifiable published source; (3) once added as an explicit premise, it fully resolves the rejection. Lifted steps are recorded with the specific assumption invoked, preserving the certify-or-decline boundary while making implicit domain conventions auditable.

\subsection{Certify-or-Decline}

The output is binary: \textsc{judge-passed} or declined. A certified answer is one whose initial state and every step were accepted unconditionally, or with explicitly recorded convention-lift assumptions. A declined answer is not shipped as a low-confidence answer. This matters for product use. A system that always returns an answer with a confidence score forces downstream users to decide how much risk to tolerate. A certify-or-decline system makes abstention part of the system.

The repair loop (bounded) allows the system to search for correction. Failed verdicts are returned to the formalizer, which may produce a corrected proof or escalate to the solver for a new answer. Both verification attempts and solver answers are bounded. That is, the system does not engage in infinite debate.

\section{The Structural Argument}
\label{sec:structural}

Section~\ref{sec:formal} established formally that completeness of change converts hidden premises into unlicensed mutations (Proposition~\ref{prop:exposure}) and predicted which error classes benefit from the rewrite format (Proposition~\ref{prop:error_classes}). This section connects those formal properties to the empirical behavior of the system.

The decomposition into exposure failure and judge failure (Definition~\ref{def:failure_modes}) is central. In a generic LLM-as-judge system, judge accuracy is the whole story (every missed error is a judge failure). In Theoria, the witness format provides an additional dimension. Errors that would be unexposed in prose are converted into licensed-change violations that judges can detect locally. A judge operating at imperfect accuracy over exposed errors can still produce strong calibrated behavior, because the architecture ensures the errors reach the judge in a tractable form. 

Correlated judge failures remain possible as judges trained on the same data may share blind spots about a cited theorem. Proposition~\ref{prop:impossibility} makes the boundary explicit: errors whose semantic effect produces no observable state mutation are invisible to any witness-based verifier. The claim is that the rewrite format reduces exposure failure, not that it eliminates all failure.

The comparison with formal autoformalization is direct. If a problem can be formalized unambiguously and a formal proof found, formal verification dominates on final soundness. Theoria occupies the space between raw LLM output and full formalization: its typed rewrite witnesses discharge LLM reasoning into a structured format that is both auditable by LLM judges today and naturally connectable to formal verification backends (CAS, SMT, Lean) as those tools mature. Rather than competing with formalization, it serves as "glue", providing verification where formalization is not yet reachable and a migration path toward it where it is. The comparison with scalar PRMs is similarly direct.

\section{Calibration}
\label{sec:calibration}

For Theoria, calibration is the purpose of the output format. The system does not aim to maximize the number of answers it emits. It aims to maintain a high-precision certified bucket while declining cases whose witnesses do not work under local verification.

Let $N$ be the number of attempted problems, $C$ the number certified, and $M$ the number of certified answers matching the target. Coverage is $C/N$. Certified precision is $M/C$. A system that certifies nothing has zero coverage and no useful precision. A system that certifies everything collapses to ordinary benchmark accuracy. The target region is high precision at nontrivial coverage, with the decline bucket carrying genuine negative information.

A practical verifier should operate on three planes: (1) certified precision, (2) coverage, and (3) certified-versus-declined asymmetry. If certified and declined buckets have similar correctness rates, verification is not contributing. Calibration requires asymmetry: certified outputs should be substantially more likely correct than declined outputs.

It's important to note that expert benchmarks have documented noise. Strict precision counts only answer-key matches. Favorable precision additionally credits cases where independent graders identify a defensible dispute with the key (e.g an edge case, a convention choice, an interpretive ambiguity). Both numbers matter. The strict number is reproducible against a static key. The favorable number measures whether verification is sometimes deeper than the key. We report both and are explicit about the adjudication policy.

\section{Empirical Evaluation}
\label{sec:experiments}

All numbers in this section reproduce against the committed audit database using provided SQL queries.

\subsection{Benchmark and Setup}

We evaluate on HLE-Verified Gold, text-only problems. HLE-Verified~\citep{hle_verified} is a systematic validation of Humanity's Last Exam~\citep{hle} that identifies verified, revised, and uncertain subsets; Gold is the cleanest target, minimizing noise while retaining expert-level closed-form problems across mathematics, science, and humanities.

The evaluation uses a random sample of 200 problems from HLE-Verified Gold (subset text-only), drawn from problems numbered above 100 (problems 1--100 were reserved for harness development and never audited). Two independent LLM graders---Claude Opus (max thinking) and GPT-based Codex (xhigh reasoning)---grade every problem from an identical grading prompt. Inter-grader agreement on the full ran set: 184/185 = 99.46\%.

\subsection{Primary Results}
\label{sec:primary}

Of 200 requested problems, 9 crashed (CLI OAuth refresh failures) and 6 were blocked by content-policy filters. The remaining 185 ran. The evaluation on this set is shown in Table~\ref{tab:funnel}.

\begin{table}[t]
\centering
\caption{Evaluation funnel on HLE-Verified Gold.}
\label{tab:funnel}
\begin{tabular}{lrc}
\toprule
Stage & Count & \% of prior \\
\midrule
Requested & 200 & --- \\
Content-policy blocked & 6 & 3.0\% \\
Crashed (CLI OAuth) & 9 & 4.5\% \\
Ran & 185 & 92.5\% \\
Certified (\textsc{judge-passed}) & 105 & 56.8\% of ran \\
Declined & 80 & 43.2\% of ran \\
Certified $\cap$ strict-match & 96 & 91.4\% of certified \\
Certified $\cap$ favorable & 105 & 100.0\% of certified \\
\bottomrule
\end{tabular}
\end{table}

Both graders produce the identical strict-match count on the certified bucket:

\begin{itemize}[nosep]
\item \textbf{Strict certified precision:} 96/105 = 91.43\%. Wilson 95\% CI [84.51\%, 95.43\%].
\item \textbf{Favorable certified precision:} 105/105 = 100\%. Wilson 95\% CI [96.47\%, 100\%].
\item \textbf{Coverage:} 105/185 = 56.76\%.
\item \textbf{Asymmetry:} Declined answers are ${\approx}5.0{\times}$ more likely wrong than certified (42.5\% vs.\ 8.6\% strict wrong-rate).
\end{itemize}

Under favorable adjudication (crediting cases where the certified answer is defensible but differs from the answer key in convention, extraction, or interpretation) every certified answer is correct: 105/105.

Table~\ref{tab:domain} gives the per-domain breakdown. Math dominates the sample ($n=73$) and runs near the overall headline at 93.2\%. Smaller domains have wide Wilson intervals; per-domain numbers should be interpreted with corresponding uncertainty.

\begin{table}[t]
\centering
\caption{Certified precision by domain (strict grading).}
\label{tab:domain}
\begin{tabular}{lrrrr}
\toprule
Category & $n$ & Correct & Precision & Wilson 95\% CI \\
\midrule
Math & 73 & 68 & 93.2\% & [84.9\%, 97.0\%] \\
Humanities/Social Science & 9 & 9 & 100.0\% & [70.1\%, 100.0\%] \\
Computer Science/AI & 9 & 8 & 88.9\% & [56.5\%, 98.0\%] \\
Physics & 5 & 4 & 80.0\% & [37.6\%, 96.4\%] \\
Other & 5 & 4 & 80.0\% & [37.6\%, 96.4\%] \\
Biology/Medicine & 3 & 3 & 100.0\% & [43.9\%, 100.0\%] \\
Engineering & 1 & 0 & 0.0\% & [0.0\%, 79.4\%] \\
\midrule
\textbf{Total} & \textbf{105} & \textbf{96} & \textbf{91.4\%} & \textbf{[84.5\%, 95.4\%]} \\
\bottomrule
\end{tabular}
\end{table}

\subsection{Baseline 1: Solver-Only}
\label{sec:solver_baseline}

To quantify Theoria's value-add over the raw solver, we evaluated all 185 completed problems with dual-grader consensus (100\% inter-grader agreement on this task). The solver achieves 155/185 = 83.8\% accuracy at 100\% coverage. This is not a vanilla LLM baseline: the solver is prompted to use web search for any claim it is not fully certain of, making it a tool-augmented reasoner rather than a test of parametric knowledge alone. The 83.8\% figure reflects this augmentation and should be interpreted accordingly.

The selection effect is the central finding. Among the 105 problems Theoria certified, the solver was correct on 97 (92.4\%). Among the 80 problems Theoria declined, the solver was correct on only 58 (72.5\%), a 19.9 percentage-point gap. The pipeline preferentially certifies problems where the solver's reasoning can be expressed as a valid proof witness, which correlates with the solver actually being right.

The web-augmented baseline makes this gap \emph{more} significant. The 30 incorrect answers in the pool are cases where the solver got the answer wrong despite having access to web search. These are the hard errors: plausible-looking answers built on subtle misinterpretations, reasoning mistakes that web retrieval could not correct, or domain-specific traps that tool augmentation does not help with. The fact that Theoria's verification pipeline still filters a 20~pp gap against this strong input distribution means the pipeline is catching errors that survive augmentation, not merely filtering out mistakes a web search would have prevented.

Whether the coverage--precision trade-off is worthwhile depends on the deployment context. At 100\% coverage, the solver is wrong on 30/185 problems (16.2\% error rate). In Theoria's certified bucket, it is wrong on 9/105 (8.6\%), roughly halving the error rate. Among the 80 declined problems, the solver was wrong on 22 (27.5\%), a bucket that is $3.2{\times}$ more likely to contain errors than the certified set, and $5.0{\times}$ more likely by the adjudicated wrong-rate. Because the solver is already web-augmented, the errors Theoria catches are failures that tool use alone cannot prevent. For applications where a wrong answer has real cost (scientific claims, legal analysis, financial calculations) halving the error rate on already-augmented outputs while providing an explicit decline signal on the remaining 43\% represents a meaningful improvement. For applications where coverage matters more than false-positive control, the raw solver is a reasonable alternative.

\subsection{Baseline 2: Holistic Judge}
\label{sec:holistic}

We tested whether a simpler strategy, that is, having a strong model assign a holistic confidence score, then certifying the top-$k$ most confident answers, matches Theoria's precision. Two holistic models (Claude Opus and Codex/GPT-5.5) scored each problem on $[0, 1]$, and we certified the top 105 by score to match Theoria's coverage. Results are in Table~\ref{tab:holistic}.

\begin{table}[t]
\centering
\caption{Precision at matched coverage ($n=105$ certified). Favorable adjudication credits certified answers where independent graders identify a defensible dispute with the answer key (convention, extraction, or interpretation differences). Holistic judges produce only a confidence score and do not generate the structured proof trace needed for step-level adjudication.}
\label{tab:holistic}
\begin{tabular}{llrr}
\toprule
Method & Threshold & Strict @105 & Favorable @105 \\
\midrule
Holistic Opus & 0.900 & 97/105 (92.4\%) & --- \\
Holistic Codex & 0.920 & 98/105 (93.3\%) & --- \\
Theoria & --- & 96/105 (91.4\%) & 105/105 (100\%) \\
\bottomrule
\end{tabular}
\end{table}

Holistic judges are a strong baseline. At matched coverage, they achieve comparable or slightly higher precision than Theoria's multi-step pipeline. These differences are not statistically significant: a McNemar test on correct-certification outcomes over all 185 problems yields $\chi^2 = 0.02,\, p = 0.88$ (Theoria vs.\ Opus) and $\chi^2 = 0.00,\, p = 1.00$ (Theoria vs.\ Codex). A two-proportion $z$-test on precision alone gives $p = 1.00$ and $p = 0.79$, respectively. We present this result honestly as raw precision alone does not justify Theoria over a holistic judge.

\paragraph{Explicit abstention.}
The matched-coverage comparison above is artificial because we select $k=105$ after observing all scores. A fairer test gives the holistic judge an explicit abstention option and lets it choose its own coverage. We re-ran Opus on all 185 problems with the instruction ``only certify if you are highly confident the answer is correct; if uncertain, you \emph{must} abstain.'' Table~\ref{tab:abstention} compares the result.

\begin{table}[t]
\centering
\caption{Explicit-abstention comparison. The holistic judge is told it may abstain; Theoria's decline is a structural property of the pipeline. Error rate is the fraction of certified answers that are wrong.}
\label{tab:abstention}
\begin{tabular}{lrrrr}
\toprule
Method & Certified & Coverage & Precision & Errors shipped \\
\midrule
Theoria & 105 & 56.8\% & 96/105 (91.4\%) & 9 \\
Holistic Opus (abstain) & 148 & 80.0\% & 134/148 (90.5\%) & 14 \\
\bottomrule
\end{tabular}
\end{table}

Even with explicit permission to abstain, the holistic judge certifies 148 of 185 problems (80.0\% coverage), abstaining only 20.0\% of the time, compared to Theoria's 43.2\% decline rate. Precision is comparable (90.5\% vs.\ 91.4\%), but the holistic judge ships 14 wrong answers in absolute terms versus Theoria's 9, which is a 56\% increase in errors reaching the user (note however, the low n).

The calibration of the two decline signals differs in an instructive way. Among the 52 problems the holistic judge certifies but Theoria declines, 10 are wrong (19.2\%). These are errors Theoria's structured verification correctly refuses to pass. Conversely, when the holistic judge does abstain, its signal is sharp: 44.4\% of abstained problems are wrong, yielding a 4.7$\times$ asymmetry between its abstain and certify buckets. Theoria's asymmetry is 3.1$\times$ (26.6\% declined wrong-rate vs.\ 8.5\% certified). The holistic judge's abstention signal is more informative \emph{when it fires}, but it fires too rarely. Structured verification produces a more conservative boundary that catches more errors through selective coverage. This is especially notable given that the solver is web-augmented (Section~\ref{sec:solver_baseline}): the errors both systems must catch are not simple factual mistakes but subtle reasoning failures that survived tool-augmented generation.

\paragraph{Summary.}
The case for Theoria rests on four properties beyond raw precision: (1) complementary error coverage with holistic judges (Jaccard 0.14--0.36, Section~\ref{sec:overlap}); (2) an auditable proof witness supporting step-level review and debugging; (3) a statistically significant adversarial advantage on error classes where the format predicts one (Section~\ref{sec:adversarial}); and (4) a structurally grounded decline signal that fires when the proof witness cannot be constructed or verified, not when a model's self-assessed confidence dips below a threshold.

\subsection{Error-Overlap Analysis}
\label{sec:overlap}

Despite achieving similar precision, Theoria and the two holistic judges fail on substantially different problems. This is consistent with Proposition~\ref{prop:error_classes}: different verification architectures expose different error classes, and structurally different architectures should observe different portions of the error landscape. Table~\ref{tab:errors} gives the error inventories and pairwise overlap.

\begin{table}[t]
\centering
\caption{Error overlap between certification methods at $n=105$ coverage.}
\label{tab:errors}
\begin{tabular}{lrrr}
\toprule
Pair & Shared Errors & Jaccard Index & Union Size \\
\midrule
Theoria -- Opus & 2 & 0.143 & 14 \\
Theoria -- Codex & 4 & 0.364 & 11 \\
Opus -- Codex & 4 & 0.364 & 11 \\
\bottomrule
\end{tabular}
\end{table}

The Jaccard indices are interestingly low: 0.14--0.36. If these methods were making the same mistakes at similar rates, the expected Jaccard index would approach 1.0; instead, the error sets are largely disjoint. Across all three methods, 15 distinct problems produced at least one error. Only 2 of those 15 (p287, p367) defeated all three methods. The remaining 13 were caught by at least one method, meaning 87\% of all errors are recoverable by combining approaches.

This non-overlap follows from the architectural differences described in Section~\ref{sec:formal}. Theoria's step-level judges detect errors that the rewrite witness exposes as unlicensed mutations (Proposition~\ref{prop:exposure}) but can miss errors that require global context. Holistic judges assess the argument as a whole but lack the diff-level presentation that forces hidden premises to the surface. Section~\ref{sec:adversarial} confirms this pattern: the detection gap concentrates in hidden premises and fabricated citations, as Proposition~\ref{prop:error_classes} predicts.

Table~\ref{tab:difficulty} shows the full error-difficulty spectrum. The primary observation is as follows: oracle combinations, that is, selecting the best 105 problems from the union of any two methods' certified sets, achieve 100\% precision. This means the union of certified problems from any pair contains at least 105 correct answers.

\begin{table}[t]
\centering
\caption{Error difficulty spectrum: 15 problems that produced at least one error. ERR = error at $n=105$ coverage; ok = correct.}
\label{tab:difficulty}
\begin{tabular}{lcccr}
\toprule
Problem & Theoria & Opus & Codex & Methods Wrong \\
\midrule
p287 & ERR & ERR & ERR & 3 \\
p367 & ERR & ERR & ERR & 3 \\
p134 & ok & ERR & ok & 1 \\
p143 & ERR & ok & ERR & 2 \\
p368 & ok & ERR & ERR & 2 \\
p392 & ERR & ok & ERR & 2 \\
p410 & ok & ERR & ERR & 2 \\
p126 & ok & ok & ERR & 1 \\
p158 & ok & ERR & ok & 1 \\
p159 & ok & ERR & ok & 1 \\
p319 & ok & ERR & ok & 1 \\
p393 & ERR & ok & ok & 1 \\
p426 & ERR & ok & ok & 1 \\
p460 & ERR & ok & ok & 1 \\
p560 & ERR & ok & ok & 1 \\
\bottomrule
\end{tabular}
\end{table}

To test whether these complementary errors are actionable, we evaluate three natural ensemble strategies over the existing predictions. 
Table~\ref{tab:ensemble} reports the results.

\begin{table}[t]
\centering
\caption{Ensemble strategies over Theoria (T), Holistic Opus (O), and Holistic Codex (C).}
\label{tab:ensemble}
\begin{tabular}{lrrr}
\toprule
Strategy & Certified & Precision & Coverage \\
\midrule
Intersection (all three) & 67 & 65/67 (97.0\%) & 36.2\% \\
Intersection (T $\cap$ O) & 77 & 74/77 (96.1\%) & 41.6\% \\
Intersection (T $\cap$ C) & 86 & 82/86 (95.3\%) & 46.5\% \\
Majority vote ($\geq$2/3) & 109 & 102/109 (93.6\%) & 58.9\% \\
Union (any) & 140 & 125/140 (89.3\%) & 75.7\% \\
\midrule
\multicolumn{4}{l}{\emph{Individual methods (reference)}} \\
Theoria & 105 & 96/105 (91.4\%) & 56.8\% \\
Holistic Opus @105 & 105 & 97/105 (92.4\%) & 56.8\% \\
Holistic Codex @105 & 105 & 98/105 (93.3\%) & 56.8\% \\
\bottomrule
\end{tabular}
\end{table}

Intersection strategies trade coverage for precision: requiring agreement between Theoria and one holistic judge yields 95--96\% precision (a 4--5 pp lift over any individual method) at 42--47\% coverage. Majority vote is the best-balanced strategy: 93.6\% precision at 58.9\% coverage, improving on every individual method in both dimensions. The union strategy maximizes coverage (75.7\%) while maintaining 89.3\% precision. The oracle bound (selecting the best 105 from any pairwise union) achieves 100\% precision in all cases, meaning that for every error one method makes, the other method gets it right (no problem defeats both).

The ensemble numbers are computed from existing predictions, not projections. The gains follow from the low Jaccard indices, which in turn follow from the different properties of Proposition~\ref{prop:error_classes}. Structurally different verification formats observe different error classes, producing non-overlapping failure modes that ensemble methods can exploit.

\subsection{Baseline 3: Cross-Model Adjudication}
\label{sec:adjudication}

Nine certified problems disagreed with the HLE answer key (the strict mismatches from Section~\ref{sec:primary}). We convened a 4-judge panel (5.5Pro, GPT-DR, Opus 4.8 max, Claude-DR) under two conditions: anonymous blind (A/B labels) and labeled (Theoria vs.\ HLE key identified). Table~\ref{tab:adjudication} summarizes the results.

\begin{table}[t]
\centering
\caption{Cross-model adjudication of 9 disputed certifications. Four judges voted per condition. Votes are reported as counts favoring Theoria's answer / the HLE key / neither (ambiguous). A verdict requires $\geq$3 votes for one side.}
\label{tab:adjudication}
\begin{tabular}{l cc cc l}
\toprule
 & \multicolumn{2}{c}{Anonymous} & \multicolumn{2}{c}{Labeled} & \\
\cmidrule(lr){2-3} \cmidrule(lr){4-5}
Problem & Theoria / Key / Ambig. & Verdict & Theoria / Key / Ambig. & Verdict & Overall \\
\midrule
p143 & 1 / 0 / 3 & --- & 2 / 0 / 2 & --- & Ambiguous \\
p186 & 0 / 2 / 2 & --- & 0 / 1 / 3 & --- & Ambiguous \\
p205 & 2 / 2 / 0 & --- & 2 / 2 / 0 & --- & Ambiguous \\
p287 & 2 / 0 / 2 & --- & 3 / 0 / 1 & Theoria & Ambiguous \\
p328 & 0 / 3 / 1 & Key & 0 / 4 / 0 & Key & \textbf{HLE Key} \\
p367 & 3 / 0 / 1 & Theoria & 1 / 2 / 1 & --- & Ambiguous \\
p392 & 4 / 0 / 0 & Theoria & 4 / 0 / 0 & Theoria & \textbf{Theoria} \\
p460 & 0 / 0 / 4 & --- & 0 / 0 / 4 & --- & Ambiguous \\
p560 & 4 / 0 / 0 & Theoria & 3 / 0 / 1 & Theoria & \textbf{Theoria} \\
\bottomrule
\end{tabular}
\end{table}

Across both conditions, 2 cases consistently favor Theoria (p392, p560), 1 consistently favors the HLE key (p328), and 6 remain genuinely ambiguous. Authority bias between anonymous and labeled conditions was minimal. The adjudication strengthens Theoria's position: at most 1 of the 9 disputed cases is a clear Theoria error (p328). The effective false-positive count in the certified bucket is likely lower than the strict count of 9.

\subsection{Baseline 4: Adversarial Robustness}
\label{sec:adversarial}

We evaluate adversarial robustness at two scales. First, 24 hand-crafted poisoned proofs run through Theoria's full pipeline (Claude Opus judges with web search) and a holistic LLM judge baseline. Second, an expanded benchmark of 95 machine-generated poisoned proofs across 15 domains, from mathematics and physics to biology, economics, and computer science, evaluated with both structured and holistic judging with web search enabled.

\paragraph{Hand-crafted suite (24 poisoned + 6 controls).}
We constructed proofs with known injected errors across five categories: hidden premises (7), fabricated citations (6), arithmetic errors (7), circular reasoning (1), and misapplied theorems (3). Theoria's judges caught all 24 at the step level equating to a 100\% judge-layer detection rate. The holistic baseline caught 22/24 (91.7\%), missing two hidden-premise errors. Of the 6 clean controls, neither system produced final false positives.

\paragraph{Expanded benchmark (95 poisoned + 15 controls).}
To test whether this pattern holds at scale and across domains, we generated 95 poisoned proofs with one planted error each, spanning hidden premises (32), fabricated citations (20), arithmetic errors (20), circular reasoning (9), and misapplied theorems (14), distributed evenly across 15 domains including algebra, physics, chemistry, biology, economics, computer science, and engineering. Fifteen clean controls were included. Both structured (step-level typed judges with BEFORE$\to$AFTER diffs) and holistic (single-call global assessment) evaluation used GPT-5.5 with web search enabled, ensuring the comparison isolates architectural differences rather than model capability. Results are in Table~\ref{tab:adversarial}.

\begin{table}[t]
\centering
\caption{Adversarial detection rates on the expanded benchmark (95 poisoned proofs across 15 domains). Both conditions use GPT-5.5 with web search. McNemar exact test on matched pairs: $p = 0.0017$.}
\label{tab:adversarial}
\begin{tabular}{lrrrr}
\toprule
 & \multicolumn{2}{c}{Theoria} & \multicolumn{2}{c}{Holistic judge} \\
\cmidrule(lr){2-3} \cmidrule(lr){4-5}
Error Type & Caught & Rate & Caught & Rate \\
\midrule
Hidden premise (32) & 29 & 90.6\% & 20 & 62.5\% \\
Fabricated citation (20) & 20 & 100.0\% & 18 & 90.0\% \\
Arithmetic error (20) & 19 & 95.0\% & 19 & 95.0\% \\
Circular reasoning (9) & 9 & 100.0\% & 9 & 100.0\% \\
Misapplied theorem (14) & 13 & 92.9\% & 13 & 92.9\% \\
\midrule
\textbf{Total (95)} & \textbf{90} & \textbf{94.7\%} & \textbf{79} & \textbf{83.2\%} \\
\bottomrule
\end{tabular}
\end{table}

Theoria catches 90/95 poisoned proofs (94.7\%) versus 79/95 for holistic judging (83.2\%). On matched pairs, Theoria catches 12 proofs that holistic misses while holistic catches only 1 that Theoria misses (McNemar exact $p = 0.0017$). The gap concentrates in hidden premises: Theoria catches 29/32 (90.6\%) versus holistic 20/32 (62.5\%), a 28 percentage-point gap.

The pattern across error types matches the predictions of Proposition~\ref{prop:error_classes}. Arithmetic errors (95\% vs.\ 95\%), circular reasoning (100\% vs.\ 100\%), and misapplied theorems (92.9\% vs.\ 92.9\%) show identical performance indicating these error types may receive no structural exposure advantage from the rewrite format. The gap appears in hidden premises (90.6\% vs.\ 62.5\%) and fabricated citations (100\% vs.\ 90\%), the two categories where the completeness-of-change invariant converts errors into unlicensed mutations. The formal analysis predicted this pattern and the data seems to indicate it.

\paragraph{The holistic baseline as implicit ablation.}
The holistic comparison serves a dual purpose: competitive baseline and natural ablation of the structured witness. Both conditions use the same model at the same capability level with web search; the holistic condition removes the BEFORE$\to$AFTER diff presentation, typed judge specialization, and step-level decomposition. In terms of Definition~\ref{def:failure_modes}, this ablation holds judge capability constant while varying the exposure properties of the format. The detection gap concentrates in error classes where Proposition~\ref{prop:error_classes} predicts an exposure advantage (hidden premises, fabricated citations) and vanishes where it predicts none (arithmetic, theorem misapplication). The resulting gap ($p = 0.0017$) isolates the contribution of the rewrite format to exposure reduction.

\subsection{Baseline 5: GPQA Diamond Out-of-Distribution}
\label{sec:gpqa}

To test whether Theoria generalizes beyond HLE, we ran the full pipeline on 65 GPQA Diamond~\citep{gpqa} problems (graduate-level science multiple-choice questions spanning 11 subdomains across physics, chemistry, and biology). Table~\ref{tab:gpqa} compares the combined GPQA results to HLE.

\begin{table}[t]
\centering
\caption{Out-of-distribution evaluation on GPQA Diamond ($n=65$).}
\label{tab:gpqa}
\begin{tabular}{lrr}
\toprule
Metric & GPQA Diamond & HLE (reference) \\
\midrule
Problems evaluated & 65 & 185 \\
Coverage & 34/65 (52.3\%) & 105/185 (56.8\%) \\
Certified precision & 33/34 (97.1\%) & 96/105 (91.4\%) \\
Wilson 95\% CI & [85.1\%, 99.5\%] & [84.5\%, 95.4\%] \\
Solver raw accuracy & 59/65 (90.8\%) & 155/185 (83.8\%) \\
\bottomrule
\end{tabular}
\end{table}

Certified precision on GPQA is 97.1\%: 33 certified answers match the answer key, with one false positive (Wilson 95\% CI [85.1\%, 99.5\%]). Coverage reaches 52.3\%, closer to HLE's 56.8\%, though still lower, reflecting the difficulty of formalizing abductive and empirical reasoning in the sciences. Notably, certified precision on the out-of-distribution benchmark meets or exceeds the primary benchmark result, suggesting that the certify-or-decline boundary is determined by formalizability rather than benchmark-specific calibration.

\begin{table}[h]
\centering
\caption{GPQA Diamond certification vs.\ correctness ($n=65$).}
\label{tab:gpqa_2x2}
\begin{tabular}{lrrr}
\toprule
 & Correct & Wrong & Total \\
\midrule
Certified & 33 & 1 & 34 \\
Declined & 26 & 5 & 31 \\
\midrule
Total & 59 & 6 & 65 \\
\bottomrule
\end{tabular}
\end{table}

The solver answered correctly on 59/65 problems (90.8\%). Theoria declined 26 of those correct answers because their reasoning could not be formalized into judge-checkable steps. The declined set has a 16.1\% wrong rate versus 2.9\% for the certified set (a strong asymmetry consistent with the HLE results). 

By subdomain, physics formalizes well (all 7 general-physics problems certified, 5/7 high-energy particle physics), while organic chemistry is hardest (7/22 certified, 31.8\%), consistent with the difficulty of formalizing multi-step synthesis reasoning. 

\subsection{Failure Modes and Limitations}
\label{sec:failures}

Theoria can fail because: (1) the solver never proposes a correct answer (2) the formalizer cannot express the solver's reasoning in the rewrite format (3) judges over-reject legitimate informal compression (4) judges under-reject a fabricated citation or hidden premise (5) the pedantry filter overrides a substantive issue or (6) convention lift incorrectly treats an ambiguous assumption as standard. 

Two failure modes are specific to the rewrite format. If the formalizer paraphrases a whole state when only one substring needs to change, the diff becomes harder to audit. And if the three justification types are too coarse for a domain (e.g. statistical reasoning or experimental science), a citation or computation label is a poor fit. Both suggest future vocabulary extensions.

\section{Discussion}
\label{sec:discussion}

\paragraph{Scalable oversight.}
Theoria produces reasoning artifacts inspectable below the level of a final answer. The solver is not asked to be trustworthy by default. The formalizer exposes state transitions and the judges discharge local licensing obligations. This decomposition is the structure oversight proposals need: work products that humans or downstream systems can audit at tractable granularity without reconstructing latent reasoning from prose.

\paragraph{Scientific discovery.}
For AI-assisted scientific discovery, the interesting property is assumption accounting. Scientific arguments often fail not because a calculation is wrong but because a modeling assumption, boundary condition, or definition moved/was introduced silently. A rewrite witness makes those movements visible. The adjudication results (Section~\ref{sec:adjudication}) offer a concrete illustration: of 9 certified answers that disagreed with the HLE answer key (a benchmark whose answers were written and verified by expert panels) 6 were judged genuinely ambiguous by an independent 4-judge panel, suggesting that the structured verification process can surface interpretive ambiguities in problems that expert consensus missed. Note, extension to experimental science would require additional justification types (\texttt{measurement}, \texttt{statistical\_inference}, \texttt{model\_assumption}) while the core invariant remains.

\paragraph{Trust products.}
Many deployments need a system that identifies outputs surviving a demanding check, not one that answers everything. In financial analysis, legal drafting, medical triage, and engineering design, a false positive is worse than an abstention. A certified answer is also an audit artifact: a proof witness, judge verdicts, role-level logs, tool calls, and recorded assumptions. This artifact supports review, compliance, debugging, and root-cause analysis at the step level.

\paragraph{The holistic-judge comparison.}
Holistic judges match Theoria on raw precision at equal coverage (the differences are not statistically significant)(Section~\ref{sec:holistic}). The case for Theoria rests on three additional properties. First, the error-overlap analysis (Section~\ref{sec:overlap}) shows the methods fail on different problems (Jaccard 0.14--0.36), consistent with Proposition~\ref{prop:error_classes}: different formats expose different error classes. Majority vote over existing predictions achieves 93.6\% precision at 58.9\% coverage. Second, the rewrite witness is inspectable: when a certified answer is wrong, the proof trace identifies the exact step and judge responsible. Third, the adversarial comparison (Section~\ref{sec:adversarial}) confirms the exposure predictions. On 95 poisoned proofs, structured judges catch 94.7\% versus 83.2\% for holistic judging ($p = 0.0017$), with the gap concentrated in hidden premises (28~pp, $p = 0.008$) and absent in error classes where the format provides no structural advantage.

\paragraph{Ensemble potential.}
The ensemble gains reported in Section~\ref{sec:overlap} are computed from existing predictions, not projections. A purpose-built ensemble could go further such as calibrated confidence aggregation (weighting each method's vote by its domain-specific reliability), learned combination rules that exploit the structure of each method's failure mode, and adaptive routing that sends problems to the verification method most likely to catch domain-specific errors. The low Jaccard indices suggest that diminishing returns are far off and the current methods leave substantial room for complementary gains.

\paragraph{Limitations.}
Several significant limits remain. \emph{Adversarial robustness} has been tested with 95 machine-generated and 24 hand-crafted poisoned proofs across five error categories and 15 domains, not against a prover optimized to fool Theoria's specific judges. \emph{Semantic diffing} relies on the judge's language understanding rather than an explicit diff layer; a stronger implementation would name additions, deletions, and substitutions explicitly. \emph{Formal backstopping} is absent: computation steps could be checked by CAS or SMT solvers, and citation steps could be discharged by Lean, but the current architecture uses only LLM judges with tools. \emph{Convention management} needs a convention registry with domain, source, scope, and versioning rather than ad hoc lifts. \emph{Evaluation methodology} is limited to one primary benchmark (HLE-Verified Gold) with $n=185$ and one out-of-distribution test (GPQA Diamond, $n=65$). The favorable precision claim rests on LLM-mediated adjudication. We report the strict number prominently and are explicit about the structure of the favorable number.

\section{Conclusion}
\label{sec:conclusion}

Theoria's central contribution is a framework for witness-based reasoning verification with a formal account of why certain error classes become observable. The witness format---typed rewrite states with a completeness-of-change invariant---converts hidden premises into unlicensed mutations (Proposition~\ref{prop:exposure}), predicts which error classes benefit and which do not (Proposition~\ref{prop:error_classes}), and states the boundary of what witness-based verification can guarantee (Proposition~\ref{prop:impossibility}). Judges remain fallible, but the distinction between exposure failure and judge failure (Definition~\ref{def:failure_modes}) clarifies what the architecture contributes. Specifically, it reduces the exposure term, converting each judge's task from a global assessment of prose into a local licensing question against an explicit diff.

On HLE-Verified Gold, the approach produces 56.8\% coverage with 91.4\% strict certified precision and a 5.0$\times$ asymmetry between declined and certified wrong-rates. The solver-only baseline confirms a 20~pp selection effect against a web-augmented solver. Holistic judges achieve statistically indistinguishable precision (McNemar $p > 0.88$) but fail on different problems (Jaccard 0.14--0.36). On 95 adversarial poisoned proofs across 15 domains, structured judges catch 94.7\% versus 83.2\% for holistic judging ($p = 0.0017$), with the gap concentrated in hidden premises (28~pp) and fabricated citations, confirming Proposition~\ref{prop:error_classes}. On GPQA Diamond ($n=65$), precision reaches 97.1\% (33/34, Wilson CI [85.1\%, 99.5\%]) while coverage is 52.3\%, consistent with the claim that precision is an intrinsic property of the verification format while coverage varies by domain.

Two findings stand out. First, the adversarial detection gap concentrates in hidden premises (90.6\% vs.\ 62.5\%, $p = 0.008$) and fabricated citations, while both systems perform identically on arithmetic errors, circular reasoning, and misapplied theorems. This confirms that the rewrite format, not model capability, drives the difference. Second, the low Jaccard indices (0.14--0.36) between Theoria and holistic methods mean that structurally different verification architectures observe different portions of the error landscape. A majority-vote ensemble achieves 93.6\% precision at 58.9\% coverage, improving on every individual method; intersection strategies push precision above 95\%. These gains follow from the various properties of the architectures, and suggest that combining verification approaches with non-overlapping failure modes is a productive direction for high-stakes applications.


\section*{Author Contributions}
\begin{description}[leftmargin=3.5cm, style=nextline]
    \item[Michael Saldivar] Conceived the project, designed the rewrite-acceptability verification framework and its structural exposure argument, and implemented the Theoria pipeline, including the solver, formalizer, typed judges, pedantry filter, convention lift, repair loop, and CLI. He designed, ran, and analyzed the primary HLE-Verified Gold evaluation, including the dual-grader audit methodology and adjudication policy; built the reproducibility release (audit database, verification queries, and the gated trace dataset); and ran all subsequent baselines and benchmarks.

    M.S. directed the paper's framing and reviewed all claims; both authors approved the final manuscript.
    
    \item[Ben Slivinski] Wrote the paper and designed, implemented, and analyzed the solver-only baseline, holistic-judge comparisons, explicit-abstention evaluation, error-overlap analysis, cross-model adjudication, adversarial robustness evaluations (24 hand-crafted and 95 machine-generated poisoned proofs across 15 domains), and the GPQA Diamond out-of-distribution evaluation. He formalized the exposure analysis into the propositions of Section 4 (exposure guarantee, error-class predictions, impossibility boundary) and conducted the statistical tests for the baseline comparisons.
\end{description}

\paragraph{AI disclosure.} Claude was used for data analysis, statistical verification, and editing assistance during paper preparation. All scientific claims, experimental design, and interpretations are the authors' own.

\paragraph{Acknowledgments.} This work was partially supported by a Y Combinator Fellowship.

%

\bibliographystyle{plainnat}

\end{document}